\documentclass{article}

\usepackage[preprint, nonatbib]{neurips_2024}

\usepackage{amsmath,amssymb,amsfonts}
\usepackage{algorithm}
\usepackage{algpseudocode}
\usepackage{graphicx}
\usepackage{textcomp}
\usepackage{xcolor}
\usepackage[english]{babel}
\usepackage[square,numbers]{natbib}
\usepackage[utf8]{inputenc} 
\usepackage[T1]{fontenc}    
\usepackage{hyperref}       
\usepackage{url}            
\usepackage{booktabs}       
\usepackage{amsfonts}       
\usepackage{nicefrac}       
\usepackage{microtype}      
\usepackage{xcolor}         

\usepackage{caption}
\usepackage{subcaption}
\usepackage{cleveref}

\title{Awareness of uncertainty in classification \\
using a multivariate model and multi-views}

\author{%
  Alexey Kornaev \\
  Research Center \\
  for Artificial Intelligence\\
  Innopolis University\\
  Innopolis, PC 420500 \\
  \texttt{a.kornaev@innopolis.ru} \\
  \And
  Elena Kornaeva \\
  Dept. of Information Systems \\
  and Digital Technologies \\
  Orel State University\\
  Orel, PC 302026 \\
  \texttt{smkornaeva@gmail.com} \\
  \AND
  Oleg Ivanov \\
  Higher School \\
  of Digital Culture \\
  ITMO University \\
  St. Petersburg, PC 197101 \\
  \texttt{411378@niuitmo.ru} \\
  \And
  Ilya Pershin \\
  Research Center \\
  for Artificial Intelligence\\
  Innopolis University\\
  Innopolis, PC 420500 \\
  \texttt{i.pershin@innopolis.ru} \\
  \And
  Danis Alukaev \\
  Research Center \\
  for Artificial Intelligence\\
  Innopolis University\\
  Innopolis, PC 420500 \\
  \texttt{d.alukaev@innopolis.university} \\
}

\begin{document}

\maketitle

\begin{abstract}
    One of the ways to make artificial intelligence more natural is to give it some room for doubt. Two main questions should be resolved in that way. First, how to train a model to estimate uncertainties of its own predictions? And then, what to do with the uncertain predictions if they appear?
    First, we proposed an uncertainty-aware negative log-likelihood loss for the case of N-dimensional multivariate normal distribution with spherical variance matrix to the solution of N-classes classification tasks. The loss is similar to the heteroscedastic regression loss. The proposed model regularizes uncertain predictions, and trains to calculate both the predictions and their uncertainty estimations. The model fits well with the label smoothing technique.
    Second, we expanded the limits of data augmentation at the training and test stages, and made the trained model to give multiple predictions for a given number of augmented versions of each test sample. Given the multi-view predictions together with their uncertainties and confidences, we proposed several methods to calculate final predictions, including mode values and bin counts with soft and hard weights. For the latter method, we formalized the model tuning task in the form of multimodal optimization with non-differentiable criteria of maximum accuracy, and applied particle swarm optimization to solve the tuning task. 
    The proposed methodology was tested using CIFAR-10 dataset with clean and noisy labels and demonstrated good results in comparison with other uncertainty estimation methods related to sample selection, co-teaching, and label smoothing. 
\end{abstract}

\section{Introduction}
Both data and models in machine learning face uncertainty. 
Uncertainty estimation can make the models more accurate and trusted \citep{kendall2017uncertainties}, but it remains a challenging problem \citep{ovadia2019trust, abdar2021review}. Classification models normally have an uncertainty estimator by default. Confidence value, that is the probability of the predicted class, is a good indicator for uncertainty estimation \citep{Pearce2021}, but sometimes it does not work well \citep{, nguyen2015deep}. Regression models can be relatively easily adopted to make predictions with their uncertainty estimation \citep{kendall2017uncertainties}. 

This work deals with a simple approach that expands regression models to the solution of classification tasks with uncertainty estimation. 

\section{Background}
Consider a model $\mathbf{f} = \mathbf{f}[\mathbf{x}, \mathbf{w}]$ parameterized with weights $\mathbf{w}$ that maps the input $\mathbf{x}$ into the output $\mathbf{y}\:'$ which in turn should be close to the ground-truth output $\mathbf{y}$ \footnote{Bold symbols are for tensors (vectors) and matrices.}. 

Suppose that it is necessary to set a conditional probability distribution $p(\mathbf{y}|\mathbf{x})$ to calculate the probability of ground-truth output $\mathbf{y}$ for a given input $\mathbf{x}$, or a distribution $p(\mathbf{y}|\mathbf{\Theta})$ for given parameters $\mathbf{\Theta}$ of the distribution, e.g. the mean $\mu$ and variance $\sigma^2$ parameterize the  normal distribution $\{\mu, \sigma^2 \} = \mathbf{\Theta}$, or a distribution $p(\mathbf{y}|\mathbf{f}[\mathbf{x}, \mathbf{w}])$ for the given model $\mathbf{f}[\mathbf{x}, \mathbf{w}]$ and input $\mathbf{x}$. The higher the probability $p(\mathbf{y}|\mathbf{f}[\mathbf{x}, \mathbf{w}])$, the closer the model $\mathbf{f}[\mathbf{x}, \mathbf{w}]$ is to the ground-truth. And this is a typical task for machine learning to find the appropriate parameters $\mathbf{\Theta} = \mathbf{f}[\mathbf{x}, \mathbf{w}]$ of the distribution for a given input $\mathbf{x}$ by means of a loss minimization $L(\mathbf{w}) \to min$.

Then for a given dataset $\{\mathbf{x}^{(i)}, \mathbf{y}^{(i)}\}$ of $m$ samples $(i = 1, ..., m)$ the machine learning task can be considered as the {\it negative log-likelihood} (NLL) criterion which declares that the probabilities of the ground-truth values $\mathbf{y}^{(i)}$ should be maximized for all the dataset samples $\mathbf{x}^{(i)}$~\citep{prince2023understanding}:

\begin{equation}
    \label{eq:NLL}
        \widetilde{\mathbf{w}} = \underset{\mathbf{w}}{\arg\min} \left[ - \sum_{i=1}^{m} \log \left[ p \left( \mathbf{y}^{(i)}|\mathbf{f}[\mathbf{x}^{(i)}, \mathbf{w}] \right) \right] \right] = \underset{\mathbf{w}}{\arg\min} \left[ L [\mathbf{w}] \right],
\end{equation}
where $\widetilde{\mathbf{w}}$ is the matrix of optimal parameters of the model. 

Taking into account the probability density function of a normal distribution, and in case the model predicts a mean value  $h^{(i)} = y^{(i)\:'}$ of the distribution and its variance $\sigma^2_{(i)}$ $\{h^{(i)}, \sigma^2_{(i)}\} = f[x^{(i)}, \mathbf{w}]$ for a given scalar input $x^{(i)}$, the unknown probability density function takes the form:

\begin{equation}
    \label{eq:UNDistr}
        p \left( \mathbf{y}^{(i)}|\mathbf{f}[\mathbf{x}^{(i)}, \mathbf{w} ] \right)= \frac{1}{\sqrt{2\pi\sigma_{(i)}^2}} e^{- \frac{\left(y^{(i)} - h^{(i)}\right)^2}{2\sigma_{(i)}^2}}.
\end{equation}
So, the NLL criterion can be formalized as follows~\citep{prince2023understanding, Goodfellow-et-al-2016, kendall2017uncertainties}:

\begin{equation}
    \label{eq:NLL_HetReg_0}
    L  =  - \sum_{i=1}^{m} \log \left[ \frac{1}{\sqrt{2\pi\sigma_{(i)}^2}} e^{- \frac{\left(y^{(i)} - h^{(i)}\right)^2}{2\sigma_{(i)}^2}} \right] \to min,
\end{equation}
then it can be averaged by the number of samples, simplified by removing an insignificant term, and represented as the {\it heteroscedastic} regression loss~\citep{prince2023understanding, kendall2017uncertainties}:

\begin{equation}
    \label{eq:NLL_HetReg_1}
    \begin{split}
        L_R  = \frac{1}{2m}\sum_{i=1}^{m} \left( \frac{1}{\sigma_{(i)}^2} \left(y^{(i)} - h^{(i)}\right)^2 + \log \sigma_{(i)}^2 \right).
    \end{split}
\end{equation}

The first term in \eqref{eq:NLL_HetReg_1} can learn aleatoric or data uncertainty without any uncertainty labels by increasing the variance $\sigma_{(i)}^2$ and decreasing the term. The second term in \eqref{eq:NLL_HetReg_1} regularizes the loss and prevents the model from predicting infinite uncertainty $(\sigma_{(i)}^2 \to \inf, L \to 0)$ for all the data samples \citep{kendall2017uncertainties}.

\Cref{eq:NLL_HetReg_1} can be reduced to the {\it homoscedastic} regression loss or the {\it mean squared error} (MSE) loss in case the distribution $p$ is univariate $(\sigma_{(i)}^2 = const)$ \citep{prince2023understanding, Goodfellow-et-al-2016}. 

So, uncertainty estimation method can be naturally injected to regression models, but not to classification models which normally deal with {\it cross entropy} (CE) loss \citep{Goodfellow-et-al-2016}. A way to combine regression and classification models and propose uncertainty aware loss for classification models is presented below.

\section{Methodology}

\subsection{Uncertainty-aware loss for the multiclass classification}

Consider a model $\mathbf{f}[\mathbf{x^{(i)}}, \mathbf{w}]$ parameterized with weights $\mathbf{w}$ that maps an input $\mathbf{x^{(i)}}$ into the output and includes two elements: an $N$-dimensional vector of predictions $\mathbf{h^{(i)}}$ for the $N$ classes that should be close to the one-hot encoded ground-truth or label vector $\mathbf{y}$, and a positive scalar value $\sigma_{(i)}^2$ that estimates uncertainty of the predictions $\{ \mathbf{h^{(i)}}, \sigma_{(i)}^2 \} = f[\mathbf{x^{(i)}}, \mathbf{w}]$. 

Suppose that a conditional probability distribution $p(\mathbf{y^{(i)}}|\mathbf{x^{(i)}}) = p (\mathbf{y}^{(i)}|\mathbf{f}[\mathbf{x}^{(i)}, \mathbf{w} ])$ has a form of  the multivariate normal distribution characterized by equal variances (spherical covariances) given in N-dimensional space \citep{princeCVMLI2012}:

\begin{equation}
\label{eq:MNDistr}
    p \left( \mathbf{y}^{(i)}|\mathbf{f}[\mathbf{x}^{(i)}, \mathbf{w} ] \right)= \frac{1}{\sqrt{(2\pi\sigma_{(i)}^2)^N}} e^{- \frac{\sum_{k=1}^{N} \left(y_k^{(i)} - h_k^{(i)}\right)^2}{2\sigma_{(i)}^2}},
\end{equation}
where $y_k^{(i)}$, $h_k^{(i)}$ are the one-hot encoded ground-truth and the predicted value for the $k$\textsuperscript{th} class of the $i$\textsuperscript{th} sample, respectively. 

The multivariate normal distribution from \eqref{eq:MNDistr} can be applied to the NLL criterion from \eqref{eq:NLL_HetReg_1} to propose a novel uncertainty-aware negative log-likelihood (UANLL) loss for multiclass classification:

\begin{equation}
    \label{eq:UANLL}
        L_{C} = \frac{1}{2m}\sum_{i=1}^{m} \left(\frac{1}{\sigma_{(i)}^2} \sum_{k=1}^{N}\left(y_k^{(i)} - h_k^{(i)}\right)^2 + N \log\sigma_{(i)}^2\right).
\end{equation}

\begin{figure}[t]
\centerline{\includegraphics[width=0.24\textwidth]{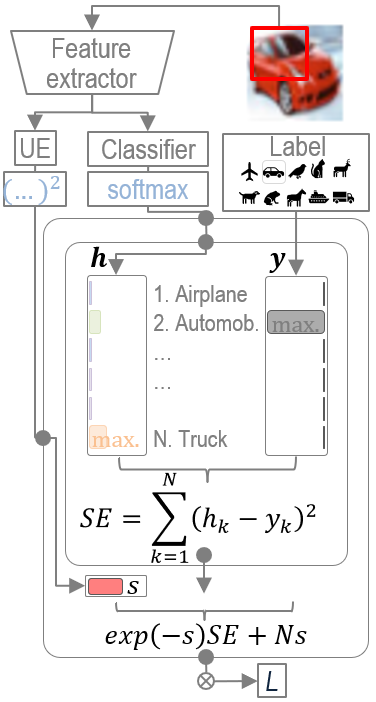}}
\caption{The proposed UANLL loss intuition: the model inputs an image sample (or its fragment) and outputs two values: a vector of predictions $\mathbf{h}$ with the components $h_k \in (0, 1)$, $\sum{h_k} = 1$, which should be close to the one-hot encoded label vector $\mathbf{y}$, and a scalar value of uncertainty estimation $s \in (0, \inf)$, and then the loss term is calculated in correspondence with \eqref{eq:UANLL_2}.}
\label{fig:LossIntuition}
\end{figure}

\citet{kendall2017uncertainties} recommended to train the NLL models to predict log variances $s^{(i)} = \log\sigma_{(i)}^2$, because it is more numerically stable than the variance $\sigma_{(i)}^2$ and the loss avoids a potential division by zero (see \Cref{fig:LossIntuition}):

\begin{equation}
    \label{eq:UANLL_2}
        L_{C} = \frac{1}{2m}\sum_{i=1}^{m} \left(e^{-s^{(i)}}  \sum_{k=1}^{N}\left(y_k^{(i)} - h_k^{(i)}\right)^2 + N s^{(i)}\right).
\end{equation}

Since the UANLL loss \eqref{eq:UANLL_2} deals with one-hot encoded labels $\mathbf{y}^{(i)}$, the label smoothing method~\citep{wei2021smooth} can be easily applied. 

\subsection{Label smoothing}
Label smoothing decreases the array component related to the true class and increases the rest of the components of one-hot encoded labels array $\mathbf{y}$ \citep{wei2021smooth}:

\begin{equation}
    \label{eq:LSintuition}
        \mathbf{y}_{LS}^{(i)} = (1 - r)\mathbf{y}^{(i)} + (r/N)\mathbf{1},
\end{equation}

where $r$ is the smooth rate $r \in [0,1]$, $N$ is the number of classes, $\mathbf{1}$ is the all one vector as $\mathbf{y}$.

\subsection{Data augmentation and multi-view predictions}
The input data augmentation allows the model to expand the dataset and decrease the effect of overfitting in cases when it may appear at the training stage. Data augmentation may also be a useful tool at the test stage when the model is already trained, especially if the model outputs predictions with their uncertainty estimations. Then some of the predictions may be uncertain. Figuratively speaking, the model may respond "I don't know" for a given input. 

One of the methods to use uncertainty estimations is to make {\it multi-view} predictions as follows. The inference model receives $n$ augmented copies of each input and calculates $n$ pairs of predicted classes and {\it certainties} of the predictions $\{ y^{(i,j)}\:', ce^{(i,j)} \}$ for each input, respectively:  

\begin{equation}
    \label{eq:pred_cer}
        y^{(i,j)}\:' =  \underset{k}{\arg\max} \left[ h_k^{(i,j)} \right], \quad ce^{(i,j)} = 1 - sigm(s^{(i,j)}),
\end{equation}
where $(i,j)$ are the dummy indexes that refer to $j$\textsuperscript{th} augmented copy of $i$\textsuperscript{th} sample, $sigm()$ is the sigmoid activation function \citep{Goodfellow-et-al-2016} that maps unlimited uncertainty estimation value into the limited interval,  $0 < ce^{(i,j)} < 1$. 

It should be noted that the maximum component of a model prediction vector $\mathbf{h}^{(i,j)}$ which is called {\it confidence} can also be interpreted as an uncertainty estimator \citep{Pearce2021}:

\begin{equation}
    \label{eq:con}
        co^{(i,j)} = \underset{k} max({h}_k^{(i,j)}),
\end{equation}
and this kind of estimators is intrinsic attribute of the majority of the classification models.  

Given the information on multi-view predictions and their uncertainty or certainty estimations, the following methods can be applied to calculate resulting predictions. 

The simplest approach is to calculate mode values over the multi-view predictions (MVM) and give final predicted classes for each $i$\textsuperscript{th} sample: 

\begin{equation}
    \label{eq:MVMode}
        y^{(i)}\:' =  \underset{j}{mode} \left[ y^{(i,j)}\:' \right].
\end{equation}

The approaches that use information on uncertainty estimation deal with cumulative weights:

\begin{equation}
    \label{eq:cum_weights}
    z_{k = y^{(i,j)}\:'}^{(i)} =  \sum_{j}{g^{(i,j)}}, \quad y^{(i)}\:' = \underset{k}{\max}{(z_k^{(i)})},
\end{equation}
where $\mathbf{z}^{(i)}$ is the additional vector with components $z_k^{(i)}$ which accumulates weights of the predicted classed over the multi-views, $g^{(i,j)}$ are the weights of predictions.  

\Cref{eq:cum_weights} represents bin count of predicted classes ${y^{(i,j)}\:'}$ with weights ${g^{(i,j)}}$. A few values can be associated with the weights, and implemented with the following methods:
\begin{itemize}
    \item soft multi-view predictions weighted with confidences $g^{(i,j)} = co^{(i,j)}$ (MVWCo-S);
    \item soft multi-view predictions weighted with certainties $g^{(i,j)} = co^{(i,j)}$ (MVWCe-S);
\end{itemize}
or their binary versions with the weights equal to zero or one:
\begin{itemize}
    \item hard multi-view predictions weighted with confidences (MVWCo-H);
    \item hard multi-view predictions weighted with certainties (MVWCe-H).
\end{itemize}

The weights binarization allows to separate multi-view predictions into two parts: the certain and uncertain predictions. One of the ways to implement the weights ${g^{(i,j)}}$ binarization is threshold tuning (see \Cref{sec:PSO}). 

\subsection{Tuning the trained model using particle swarm}
\label{sec:PSO}
The {\it particle swarm optimization} (PSO) method is able to find the global minimum of a multimodal objective function which also can be non-differentiable \citep{wang2018particle}. The main disadvantage of the PSO is necessity to set the limits of the parameters. 

Hard multi-view predictions weighted with confidences or with certainties (MVWCo-H, MVWCe-H) can be easily implemented with PSO. The main task is to find the threshold value $t$ that converts continuous values of confidences or certainties to categorical ones: $co^{(i)}, ce^{(i)} \in \{0,1\}$. The additional tasks are to find the best settings for the data augmentation, e.g. crop scale of random resized crops, if needed. \Cref{alg:PSO} represents tuning of the model that helps to find the best values of the threshold $t$ and the crop scale limits $sc$ that leads to a maximum accuracy of the model at the validation set.    

\begin{algorithm}[t]
\caption{Tuning the inference model parameters with PSO}\label{alg:PSO}
\renewcommand{\algorithmicrequire}{\textbf{Input:}}
\renewcommand{\algorithmicensure}{\textbf{Output:}}
\begin{algorithmic}[1]
\Require parameters: random resized crop scale $sc \in (0, 1]$, threshold $t \in (0, 1)$ of the weights in \eqref{eq:cum_weights}; augmentation function $aug()$, inference model $model()$, accuracy $acc()$, optimizer $pso()$ and its hyperparameters $hyp$ (number of particles, inertial weight), inputs $\bf X = [x^{(i)}]$, targets $\bf Y$; number of iterations $iter$, samples $m$ and multi-views $n$.
\Ensure tuned parameters $sc$, $t$.
\For{$p \gets 1, iter$}
\For{$i \gets 1, m$}
    \For{$j \gets 1, n$}
        \State $x^{(i,j)} \gets aug(x^{(i)},sc^{(p)})$
        \State $y^{(i,j)}\:', co^{(i,j)}, ce^{(i,j)} \gets model(x^{(i,j)})$
    \EndFor
    \State $y_{co}^{(i)}\:' \gets \cref{eq:cum_weights}, z_{k = y^{(i,j)}\:'}^{(i)} =  \sum_{j}{(co^{(i,j)} > t^{(p)})}$
    \State $y_{ce}^{(i)}\:' \gets \cref{eq:cum_weights}, z_{k = y^{(i,j)}\:'}^{(i)} =  \sum_{j}{(ce^{(i,j)} > t^{(p)})}$
\EndFor

\State $[acc_{co}, acc_{ce}] \gets [acc(\mathbf{Y}_{co}\:', \mathbf{Y}), acc(\mathbf{Y}_{ce}\:', \mathbf{Y}) ]$
\State $criterion^{(p)} \gets 1 - {\max}{[acc_{co}, acc_{ce}]}$
\State $sc^{(p)}, t^{(p)} \gets pso(criterion^{(p)},sc^{(p)}, t^{(p)},hyp)$

\EndFor
\end{algorithmic}
\end{algorithm}

\section{Related work}
\subsection{Uncertainty-aware objectives}
Uncertainty-aware predictors are concurrent with modern Bayesian approaches to deep learning \citep{sensoy2018evidential}. The predictors deal with Gaussian processes (GPs) \citep{rasmussen:williams:2006} which are characterized by the mean values (predictions) and variances (uncertainty estimations of the predictions).  
Widely used in regression {\it mean squared loss} (MSE) corresponds to the {\it homoscedastic} regression and follows naturally from the maximum likelihood estimation of the model with the assumption that the predictions are drawn from normal distributions with variable means $\mu$ and a fixed variance $\sigma^2$ \citep{prince2023understanding}. One of the approaches to uncertainty estimation of the regression models is the {\it heteroscedastic} regression that takes both the variable mean and variance into account \citep{prince2023understanding, seitzer2022pitfalls}. So, the model trains to predict means and variances, and the uncertainty of the model predictions can be estimated using the variance values. 

The work by \citet{kendall2017uncertainties} had the greatest impact on this research. The authors dealt with two types of uncertainty, that are aleatoric (data uncertainty) and epistemic (model uncertainty), and proposed two approaches in uncertainty estimation. Proposed by the authors heteroscedastic aleatoric uncertainty estimator in regression gave the key idea to this study. \citet{kendall2017uncertainties} declared that out-of-data examples cannot be identified with aleatoric uncertainty. The authors also proposed an approach that combines aleatoric and epistemic uncertainties, but this approach is out of the scope of this research.  

Fortunately, classification models can also use a {\it squared error} (SE) loss. \citet{hui2021evaluation} demonstrated that the SE and CE based computer vision models are close in accuracy. However, a SE loss needs some more training epochs. The proposed loss \eqref{eq:UANLL_2} is similar to a SE loss. 

\citet{englesson2023logisticnormal} applied {\it logistic-normal} (LN) distribution, which is defined for categorical distributions, and proposed more general loss for multiclass classification, than \eqref{eq:UANLL}. Despite the fact that the losses look similar (to each other and to the heteroscedastic regression loss \eqref{eq:NLL_HetReg_1}), they have some differences. The model proposed in \citep{englesson2023logisticnormal} can be interpreted as regression problem with a given Gaussian distribution and with targets given in a logit space, but the loss~\eqref{eq:UANLL} represents multivariate Gaussian distribution with a spherical covariances matrix and with targets given in a probability space. It should be noted that the particular case with spherical covariances was not studied in \citep{englesson2023logisticnormal}. \citet{englesson2023logisticnormal} explained in details the main limitations of the generalized approach: asymmetry of softmax centered leads to an additional (dummy) class creation, necessity to predict the components of large covariance matrices complicates training, a few additional hyperparameters which need to be tuned.    

\citet{sensoy2018evidential} developed a theory of evidence perspective and represented the model predictions as a Dirichlet density distribution over the softmax outputs and proposed a novel loss function.
\citet{collier2020simple} proposed a method for training deep classifiers under heteroscedastic label noise. The method deals with the softmax temperature tuning that allows to control a bias-variance trade-off.

\subsection{Ensembling and multi-view predictions}
Despite the fact that epistemic or model uncertainty estimation and Bayesian neural networks (BNNs) are out of the scope of this research, the widely used approach of {\it ensembling} of the networks has great influence on this work \citep{pearce2020uncertainty, ashukha2020pitfalls}. \citet{ashukha2020pitfalls} demonstrated that many ensembling techniques are equivalent to an ensemble of several independently trained networks in terms of test performance. Since this work deals with aleatoric or data uncertainty, the opposite idea of using one trained network and several augmented copies of a sample formed the idea of multi-view predictions. The idea implementation may need some model tuning, and the easiest way to implement those tuning is to use gradient free methods, e.g. particle swarm optimization \citep{shi1999empirical, wang2018particle, shami2022particle}. The term multi-view predictions came from 3D images processing \citep{dai20183dmv}.

\subsection{Data uncertainty estimation in practice}
Corrupted inputs \citep{mintun2021interaction} and corrupted labels \citep{yao2020searching}, in-domain and out-of-domain distributions \citep{Pearce2021, kendall2017uncertainties, collier2020simple} are some of the poles of the research in the scope of data uncertainty estimation. 
This work deals with corrupted labels in the in-domain distributions. The typical test of the models in practice is to use public datasets with corrupted (noisy) labels at the training and validation stages, but with the clean labels at the test stage \citep{xia2021sample,yao2020searching}. 

Probably the simplest ways to make models be more robust to noise in labels are label smoothing \citep{wei2021smooth} and data augmentation \citep{rebuffi2021data}.
\citet{han2018coteaching} declared that models learn data with clean labels first and noisy labels then, and proposed a new paradigm called {\it co-teaching} with training of two networks.   
A number of methods try to detect input samples with incorrect labels and remove \citep{collier2020simple, xia2021sample, yao2020searching} or under-weight these samples \citep{kendall2017uncertainties, englesson2023logisticnormal}. 
In terms of practical implementation and planning of experiments, the closest research is the work by \citet{xia2021sample}. 

\section{Data collection}
\label{sec:DataCollection}

The proposed methodology was tested using public dataset CIFAR-10 \citep{Krizhevsky2009LearningML} which is widely used for the evaluation of machine learning with noisy data \citep{xia2021sample,jiang2020synthetic,han2018coteaching,yao2020searching}. This paper deals with {\it asymmetric} noise in labels of 20\% and 40\% \citep{xia2021sample}. Since the classes of the corrupted labels are visually similar pairs of objects (airplane and bird, automobile and truck, cat and dog, deer and horse), this type of synthetic noise is close to real-world noise in labels. 

\section{Experiments statement}
\label{sec:ExpStatement}
Since the noise in labels is a typical challenge in uncertainty estimation problems, a number of research performed typical experiments using typical network architectures and settings.

In all the experiments the dataset was split into training, validation, and test sets in the amount of $[45000, 5000, 10000]$ samples, respectively. The test set was not corrupted with noise in labels \citep{xia2021sample}. Models were trained with randomly initialized 9-layer convolution neural network \citep{han2018coteaching, xia2021sample}. The network has 4.4 million randomly initialized trainable parameters. Most of the settings correspond to the experiments by \citet{xia2021sample}: the models were trained in 200 epochs using the Adam optimizer with momentum of $0.9$ and batch size of $128$; initial learning rate of $0.001$ linearly decreased to zero starting from $80$\textsuperscript{th} epoch; image samples were transformed to tensors and normalized with means of $[0.9133, 0.2737, 0.2737]$ and standard deviations of $[0.1576, 0.2508, 0.2508]$. 

Some of the experiments settings might differ from the settings by \citet{xia2021sample}: weight decay of $0.1$ and data augmentation with random resized crop with the crop scale $sc \in [0.1, 1]$ allowed to decrease the effect of overfitting which is typical when the data labels are noisy; label smoothing \eqref{eq:LSintuition} with smooth rate of $0.4$ was applied; models with the lowest validation loss were used as inference ones; the test set sampling was implemented using the multi-view predictions, so that each sample was augmented $50$ times with the crop scale $sc \in [0.4, 1]$, the obtained multi-view predictions allowed to calculate final predictions using proposed MVM, MVWCo-S(H), MVWCe-S(H) methods (see \Cref{sec:preliminary_exps}). 

All the experiments were performed five times with random seeds of $[42, 0, 17, 9, 3]$, then the mean and standard deviation of experimental results were reported. Accuracy and {\it expected calibration error} (ECE) with $32$ bins were used \citep{kumar2020verified} as metric. The obtained results were not compared with the state-of-the-art results. The latter aggregate multiple techniques and complex network architectures. So, the comparison would be unfair. 

\section{Results and discussion}
Experiments studied the proposed methodology of uncertainty estimation in classification models using CIFAR-10 dataset and asymmetric noise in labels~\footnote{Submitted at https://github.com/... .}. The models based on CE loss and the proposed UANLL loss~\eqref{eq:UANLL_2} were compared with each other and with other authors' results.

\subsection{Preliminary experiments and tuning of the model settings}
\label{sec:preliminary_exps}
The main difficulty faced during the training with noisy labels was overfitting of the models. To resolve the difficulty, the Adam optimizer was tuned with a few additional experiments. The model with UANLL loss~\eqref{eq:UANLL_2} and CIFAR-10 dataset with 40\% asymmetric noise in labels was tested with a few values of the {\it weight decay} $wd = [0.02, 0.04, \mathbf{0.1}, 1.0]$, and the following test accuracy results were obtained: $acc = [67.73\%, 68.14\%, \mathbf{69.70}\%, 53.78]$. As for the data augmentation, the more the augmentations were, the farther the minimum value of the validation loss was shifted along the axis of number of iterations. So, the model got more training iterations before the overfitting had started, and the higher was the accuracy of the inference model. The validation loss minimum was not shifted in case of weight decay tuning, but the weight decay had great influence on the accuracy of the inference model.   

\begin{figure}[ht]
    \begin{center}
    \begin{subfigure}[b]{0.15\textwidth}
        \centering
        \includegraphics[width=\textwidth]{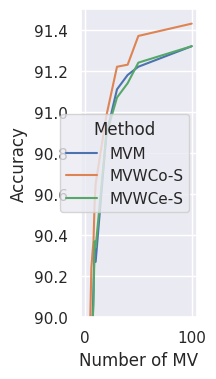} 
        \caption{}
        \label{fig:MV}
    \end{subfigure}
    \hspace{1cm} 
    \begin{subfigure}[b]{0.15\textwidth}
        \centering
        \includegraphics[width=\textwidth]{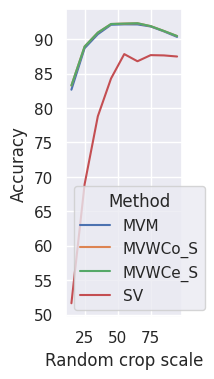} 
        \caption{}
        \label{fig:sc}
    \end{subfigure}
    \hspace{1cm} 
    \begin{subfigure}[b]{0.15\textwidth}
        \centering
        \includegraphics[width=\textwidth]{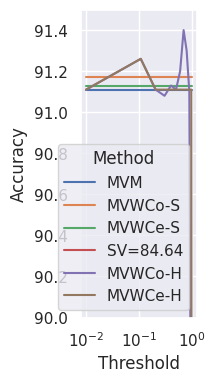} 
        \caption{}
        \label{fig:t}
    \end{subfigure}
    \end{center}
\caption{Preliminary experiments demonstrate dependence of the models accuracy on the number of multi-view predictions (a), on the random crop scale $sc$ (b), and on the threshold value for the MVWCo-H and MVWCe-H methods (c).}
\label{fig:MV_sc}
\end{figure}

Fig.~\ref{fig:MV_sc} demonstrates that the more the number of multi-view predictions, the higher the accuracy of the models.  It can be seen that the accuracy does not change much when the number of multi-view predictions is more than about $40$. The random crop scale should be equal to its maximum value of $1.0$ when the ordinary single view prediction technique is used. But in the case of multi-view predictions the maximum accuracy corresponds to the crop scale of about $0.4...0.8$. So, the crop scale should be tuned manually or using PSO method (see \Cref{alg:PSO}), as well as the threshold for the weights of hard multi-view predictions MVWCo-H and MVWCe-H. 

The generalized label smoothing with negative rates $r \in (-\inf, 1]$ proposed by \citet{wei2021smooth} did not affect the results of CE loss based models, and decreased the accuracy of the UANLL loss~\eqref{eq:UANLL_2} based models. However, the label smoothing with positive rates gave positive results. 

\subsection{The main experiments and comparison with other research}
The main experiments had the following stages. The first stage was training of the models using CE loss \citep{Goodfellow-et-al-2016} or the proposed UANLL loss~\eqref{eq:UANLL_2}. The second stage (optional) was implementation of multi-view predictions and tuning the threshold value $t$ using the validation set and PSO (see \Cref{alg:PSO}). The last stage was implementation of multi-view predictions using the test set and calculation of the accuracy and ECE of the models.

\begin{table*}[ht]
  \caption{Accuracy (\%) and expected calibration error (ECE) of the models trained with clean labels, and labels with asymmetric noise of 20\% and 40\%, and with label smoothing (LS). The best two results are in bold.}
  \centering
  \tiny
  \begin{tabular}{@{}rr ccc ccc @{}}
\hline
    {\bf No}&{\bf Method}&{\bf Arch.}&{\bf \#Param.}&{\bf LS}&\multicolumn{3}{c}{\bf Accuracy,\% / (ECE$\times10^3$)}\\
    &            &           &(train.par.)            &     & 0 & 20 & 40                      \\
\hline
    1.& Baseline CE loss&9-l.CNN&4.4 M (all)& 0.0&$87.19\pm1.03 /$&$80.06\pm0.54 /$&$65.15\pm1.99 /$ \\ 
    &\multicolumn{4}{c}{}                         &$(33.74\pm10.71)$&$(99.22\pm11.69)$&$(61.12\pm34.97)$ \\    
    &+\hspace{0.6cm} MVM & & &                    &$92.28\pm0.65$&$86.64\pm0.54$&$71.32\pm1.91$\\
    &+ MVWCo-S & & &                              &$92.34\pm0.66$&$\mathbf{86.69\pm0.56}$&$71.23\pm1.91$\\

    2.&Proposed UANLL loss&9-l.CNN&4.4 M (all)& 0.4 &$89.55\pm0.10 /$&$79.71\pm0.19 /$&$68.59\pm1.42 /$ \\
    &\multicolumn{4}{c}{}                         &$(356.6\pm1.2)$&$(377.5\pm7)$&$(333.5\pm17.3)$ \\   
    &+\hspace{0.6cm} MVM & & &                    &$93.54\pm0.08$&$86.26\pm0.25$&$75.08\pm1.25$\\
    &+ MVWCo-S           & & &                    &$\mathbf{93.61\pm0.12}$&$86.32\pm0.31$&$75.15\pm1.24$\\
    &+ MVWCe-S           & & &                    &$93.54\pm0.09$&$86.23\pm0.28$&$75.07\pm1.22$\\
    &+ MVWCo-H           & & &                    &$93.59\pm0.05$&${86.33\pm0.25}$&$\mathbf{75.20\pm1.28}$\\
    &+ MVWCe-H           & & &                    &$93.54\pm0.08$&$86.26\pm0.25$&$75.09\pm1.26$\\
    3.&CNLCU-S \citep{xia2021sample}&9-l.CNN&4.4 M (all)&$-$&$-$&$85.06\pm0.17 / -$&$\mathbf{75.34\pm0.32} / -$\\
    4.&Co-teaching \citep{xia2021sample, han2018coteaching} &9-l.CNN&4.4 M (all)&$-$&$-$&$\mathbf{86.87\pm0.24} / -$&$73.43\pm0.62 / -$\\
    5.&S2E \citep{xia2021sample, yao2020searching}&9-l.CNN&4.4 M (all)&$-$&$-$&$86.03\pm0.56 / -$&$75.08\pm1.25 / -$\\
    6.&LVCM\citep{collier2020simple}&9-l.CNN&4.4 M (all)&$-$&$89.2 / -$&$-$&$-$\\
\hline
    7.&Deep ensembles \citep{laurent2023packedensembles}&ResNet18&$-$&$-$&$\mathbf{95.1 / (8)} $&$-$&$-$\\
    8.&GMM \citep{Pearce2021}&ResNet18&$-$&$-$&$93.5\pm0.1 / -$&$-$&$-$\\
    9.&LN \citep{englesson2023logisticnormal}&WideResNet-28-2&$-$&$-$&$90.17\pm0.55 / -$&$-$&$-$\\
    10.&GLS \citep{wei2021smooth}&ResNet 34&$-$&0.4&$93.05\pm0.04 / -$&$-$&$-$\\
\hline     
  \end{tabular}
  \label{tab:RnResults}
\end{table*}

\begin{figure}[ht]
    \begin{center}
    \begin{subfigure}[b]{0.15\textwidth}
        \centering
        \includegraphics[width=\textwidth]{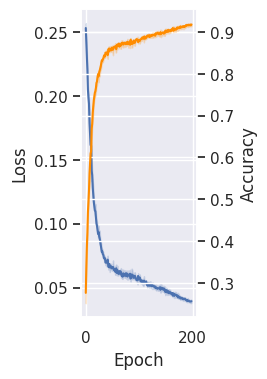} 
        \caption{0\%}
        \label{fig:UANLL_0}
    \end{subfigure}
    \hspace{1cm} 
    \begin{subfigure}[b]{0.15\textwidth}
        \centering
        \includegraphics[width=\textwidth]{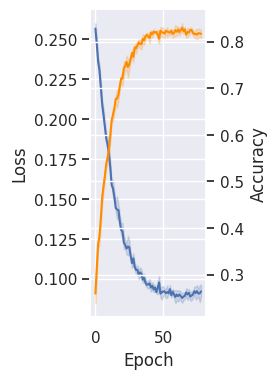} 
        \caption{20\%}
        \label{fig:UANLL_20}
    \end{subfigure}
    \hspace{1cm} 
    \begin{subfigure}[b]{0.15\textwidth}
        \centering
        \includegraphics[width=\textwidth]{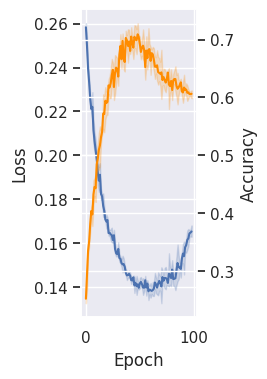} 
        \caption{40\%}
        \label{fig:UANLL_40}
    \end{subfigure}
    \end{center}
\caption{Validation loss and accuracy of the models based on the proposed loss~\eqref{eq:UANLL_2} trained with clean labels (a), and labels with asymmetric noise of 20\% (b), and 40\% (see \Cref{tab:RnResults}).}
\label{fig:Val_loss_and_acc}
\end{figure}

It should be noted that the training loss functions decreased monotonously during training in all experiments. So, it can be seen in Fig.~\ref{fig:Val_loss_and_acc} that the models based on the proposed loss~\eqref{eq:UANLL_2} demonstrated overfitting in cases of noise in labels. The same situation was with the models based on CE loss. The effect of overfitting was reduced by means of regularization of the models parameterized with weight decay $wd$ and by means of augmentation with random resized crop (see \Cref{sec:ExpStatement}). 

The obtained results in comparison with the results of other authors are presented in \Cref{tab:RnResults}. The following models were compared: the baseline model with CE loss was implemented without and with multi-view predictions; the model with the proposed UANLL loss~\eqref{eq:UANLL_2} implemented without and with multi-view predictions; method, which combats noisy labels by concerning uncertainty with soft trunication (CNLCU-S) \citep{xia2021sample}; co-teaching method \citep{han2018coteaching,xia2021sample}; search to exploit (S2E) method \citep{yao2020searching, xia2021sample}; latent variable classification model (LVCM)~\citep{collier2020simple}; Gaussian mixture model (GMM)~\citep{Pearce2021}; logistic-normal (LN) likelihood for heteroscedastic label noise \citep{englesson2023logisticnormal}; generalized label smoothing (GLS)~\citep{wei2021smooth}. It should be noted that \citet{englesson2023logisticnormal} applied asymmetric noise without one of the changes in labels (no changes in car and truck labels). So, the obtained in \citep{englesson2023logisticnormal} results with 20\% and 40\% asymmetric noise in labels cannot be compared with the other authors' results in \Cref{tab:RnResults}.    

The model based on UANLL loss~\eqref{eq:UANLL_2} got 2\textsuperscript{nd} and 3\textsuperscript{rd} places among the other 9 models in experiments with clean and noisy data. Significant difference between single-view and multi-view predictions for the models 1,2 in \Cref{tab:RnResults} is partially the result of the test data augmentation with random resized crops. 

\section{Ablation study}
The loss \eqref{eq:UANLL_2} includes an uncertainty estimator that might not have any valuable influence on the results. So, the section deals with the cropped version of the loss \eqref{eq:UANLL_2}, which is similar to the MSE loss for classification: 

\begin{equation}
    \label{eq:UANLL_abl}
        L_{A} = \frac{1}{2m}\sum_{i=1}^{m} \sum_{k=1}^{N}\left(y_k^{(i)} - h_k^{(i)}\right)^2.
\end{equation}

The experiments were performed without and with asymmetric noise in labels. The model was trained with random crop scale $sc \in [0.1, 1.0]$, weight decay of $0.1$, and with labels smoothing ratio of 0.4 in correspondence with the main experiments (see \Cref{sec:ExpStatement}, \Cref{tab:RnResults}).

The model based on the proposed UANLL loss~\eqref{eq:UANLL_2} with MVWCo-S is more accurate than its version with the cutted-off loss~\eqref{eq:UANLL_abl} at about $0.12\%$, $0.15\%$, $2.2\%$ in cases of clean labels, and labels with 20\% and 40\% of asymmetric noise, respectively. So, the uncertainty estimation in~\eqref{eq:UANLL_2} has great influence on the quality of the model (see \Cref{tab:Ablation}).

\begin{table*}[h]
\caption{Ablation study of the proposed model (see Fig.~\ref{fig:LossIntuition}, \Cref{tab:RnResults}) with the cutted-off loss~\eqref{eq:UANLL_abl}.}
  \centering
  \tiny
  \begin{tabular}{@{}r ccc @{}}
\hline
    {\bf Method}&\multicolumn{3}{c}{\bf Accuracy,\% / (ECE$\times10^3$)}\\
                 & 0 & 20 & 40                      \\
\hline
    UANLL \eqref{eq:UANLL_abl}&$89.41\pm0.18 /$&$79.95\pm0.30 /$&$66.57\pm1.15 /$\\ 
                              &$(355.9\pm1.3) $&$(377.8\pm6.6)$&$(311.3\pm19.6)$\\    
    +\hspace{0.65cm}MVM&$93.47\pm0.09  $&$86.13\pm0.50$&$72.96\pm0.77  $\\
    + MVWCo-S &$93.48\pm0.09  $&$86.16\pm0.51$&$72.95\pm0.75  $\\
    + MVWCe-S &$93.45\pm0.09  $&$86.11\pm0.50$&$72.88\pm0.72  $\\
    + MVWCo-H &$93.48\pm0.15  $&$86.01\pm0.78$&$72.58\pm0.98  $\\
    + MVWCe-H &$93.47\pm0.09  $&$86.13\pm0.50$&$72.91\pm0.80  $\\
\hline
  \end{tabular}
  \label{tab:Ablation}
\end{table*}

\section{Limitations}
The proposed loss and the corresponding models were tested with the aleatoric or data uncertainty related to the noisy labels. The proposed loss does not take into account the epistemic or model uncertainty. The models were not tested with corrupted inputs and with inputs obtained from an out-of-distribution domain. 

The proposed loss demonstrated good results. However, we could not interpret variance value which is the additional output of the classification model. We transformed the variance into the certainty value and used it in the calculation of the accuracy of the multi-view predictions. The results obtained with multi-view predictions demonstrated that a confidence value $co^{(i)}$ related to the Gaussian means of the model is more informative than a certainty value $ce^{(i)}$ related to the Gaussian variance. Also, we could not interpret high difference in terms of expected calibration error between the models based on the proposed loss and on the CE loss.

\section{Conclusion}
In most of the studied cases the proposed uncertainty-aware loss made the model more accurate in both processing data with clean and noisy labels. The proposed multi-view approach demonstrated stable positive influence on the accuracy of the models. 
The main prospects of the research are uncertainty estimation of data with corrupted inputs and application of the proposed methods of uncertainty estimation and multi-view predictions to object detection and reinforcement learning.

{
\small
\bibliography{references}
}

\end{document}